\documentclass{article}
\usepackage{bm}
\usepackage{xcolor}
\usepackage{amsthm}
\usepackage{amsmath}
\usepackage{amssymb}
\usepackage{hyperref}
\usepackage{graphicx}
\usepackage{booktabs}
\usepackage{enumitem}
\usepackage{mathtools}
\usepackage{microtype}
\usepackage{subfigure}
\usepackage[page]{appendix}
\usepackage{threeparttable}
\usepackage{multirow, makecell}
\usepackage[textsize=tiny]{todonotes}
\usepackage[capitalize,noabbrev]{cleveref}

\usepackage[accepted]{icml2024}

\begin{document}
\twocolumn[
\icmltitle{DataLight: Offline Data-Driven Traffic Signal Control}
\icmlsetsymbol{equal}{*}

\begin{icmlauthorlist}
\icmlauthor{Liang Zhang}{lzu}
\icmlauthor{Yutong Zhang}{bupt}
\icmlauthor{Jianming Deng}{lzu}
\icmlauthor{Chen Li}{ngy}
\end{icmlauthorlist}

\icmlaffiliation{lzu}{State Key Laboratory of Herbage Improvement and Grassland Agro-ecosystems, College of Ecology, Lanzhou University, Lanzhou 730000, China }
\icmlaffiliation{bupt}{School of Artificial Intelligence, Beijing University of Posts and Telecommunications}
\icmlaffiliation{ngy}{Graduate School of Informatics, Nagoya University, Chikusa, Nagoya 464-8601, Japan}

\icmlcorrespondingauthor{Jianming Deng}{dengjm@lzu.edu.cn}

\icmlkeywords{Machine Learning, ICML}

\vskip 0.3in
]
\printAffiliationsAndNotice{\icmlEqualContribution}

%%%%%%%%%%%%%%%%%%%%%%%%%%%%%%%%%%%%%%%%%%%%%%%%%%%%%%%%%%%%%%%%%%%%%%%%%%%%%%%%%%
\begin{abstract}
Reinforcement learning (RL) has emerged as a promising solution for addressing traffic signal control (TSC) challenges. While most RL-based TSC systems typically employ an online approach, facilitating frequent active interaction with the environment, learning such strategies in the real world is impractical due to safety and risk concerns. To tackle these challenges, this study introduces an innovative offline data-driven approach, called \textit{DataLight}. DataLight employs effective state representations and reward function by capturing vehicular speed information within the environment. It then segments roads to capture spatial information and further enhances the  spatially segmented state representations with sequential modeling. The experimental results demonstrate the effectiveness of DataLight, showcasing superior performance compared to both state-of-the-art online and offline TSC methods. Additionally, DataLight exhibits robust learning capabilities concerning real-world deployment issues. The code is available at \url{https://github.com/LiangZhang1996/DataLight}.
\end{abstract}

%%%%%%%%%%%%%%%%%%%%%%%%%%%%%%%%%%%%%%%%%%%%%%%%%%%%%%%%%%%%%%%%%%%%%%%%%%%%%%%%%%
\section{Introduction}
\label{sec:introduction}
% Background
Managing and alleviating traffic congestion remains a significant challenge, underscoring the pivotal role of traffic signal control (TSC) in shaping urban mobility. Traditional TSC methodologies, including FixedTime~\cite{fixedtime}, GreenWave~\cite{greenwave}, SCATS~\cite{scats}, SCOOT~\cite{scoot}, and SOTL~\cite{sotl2013}, have seen widespread adoption in recent years. However, their dependence on manually designed traffic signal plans or predefined rules limits their adaptability and flexibility in handling diverse traffic conditions. In contrast, reinforcement learning (RL)-based methods~\cite{rl} have emerged as promising solutions for adaptive TSC, owing to their capacity to dynamically adapt to evolving scenarios. RL-based models such as FRAP~\cite{frap}, CoLight~\cite{colight}, AttendLight~\cite{attend}, and Advanced-XLight~\cite{advanced} have garnered popularity in the realm of TSC, attributed to their proficiency in learning and adapting to various traffic conditions through interactions with the environment, utilizing a trial-and-error approach. 

% Online drawbacks
Most current RL-based methods typically rely on online approaches, entailing the learning of optimal policies through active interaction with the environment \cite{frap,colight,advanced}. While online RL is fundamental for adapting to dynamic scenarios and optimizing policies in real-time, it encounters challenges such as the need for extensive, active environmental interactions, which can be costly, time-consuming, and involve safety risks~\cite{BCQ,BRAC,BEAR,CQL}. Additionally, the substantial data requirement for satisfactory performance poses a hurdle in environments where data collection is expensive or challenging. As urban mobility and traffic conditions continually evolve, there is an increasing demand for offline RL-based TSC solutions to offer comprehensive insights and optimize traffic flow in a more anticipatory and strategic manner.

% Motivations
Offline RL-based approaches offer a practical solution by facilitating learning from pre-existing datasets, thus eliminating the necessity for real-time interactions and contributing to a reduction in safety and ethical concerns. The utilization of large-scale historical data proves particularly beneficial in the context of TSC, where acquiring new data can be a challenging endeavor. In the realm of offline RL, advanced methodologies include approaches such as conservative Q-learning (CQL)~\cite{CQL} and Decision Transformers~\cite{DT}. These techniques effectively transform existing datasets into potent decision-making tools, leveraging insights from a variety of static experiences. Consequently, offline RL-based approaches hold the potential to streamline the learning process, enhancing the practical applicability in various settings and promoting adaptability to complex real-world challenges.
Despite the rapid advancement of offline RL, its application in TSC remains notably scarce. Furthermore, offline RL methods consistently grapple with unresolved issues of data distribution shift, adversely affecting model scalability. Therefore, applying offline RL to TSC and deriving satisfactory models from it poses a significant challenge.

% Objective
Inspired by the above motivations, this study introduces an innovative offline data-driven approach named \textit{DataLight} for TSC, aiming to significantly enhance the efficiency of traffic management systems. Specifically, DataLight integrates a network equipped with a self-attention mechanism \cite{attention} to proficiently model spatial states, allowing for a more nuanced understanding of traffic dynamics. Furthermore, DataLight excels in capturing the intricate dynamics and spatial distributions of vehicles through the meticulous design of effective state representations and reward function. These distinctive features empower DataLight to efficiently glean insights from offline data and provide a robust framework for optimizing TSC. The main contributions are structured as follows:
\begin{itemize}[leftmargin=*]
\item \textbf{An offline data-driven RL-based model:} Unlike other RL-based TSC methods, DataLight is an offline data-driven RL approach that eliminates the need for real-time interaction in TSC systems, enhancing its practical applicability across various environments.

\item \textbf{Innovative design for state and reward function:} DataLight captures vehicle dynamics and spatial positioning by monitoring vehicle speeds and segmenting road spaces, enhancing the representation of traffic states and providing a dynamic control of the overall traffic environment.

\item \textbf{Outperforming state-of-the-art (SOTA) models:} Extensive experimental results demonstrate that DataLight outperforms all online and offline SOTA models, establishing a new benchmark for advanced TSC systems.

\item \textbf{Strong learning capabilities:} DataLight exhibits remarkable abilities in learning effective strategies from minimal amounts of offline data and cyclical offline data easily available in the real world, addressing real-world issues.
\end{itemize}

%%%%%%%%%%%%%%%%%%%%%%%%%%%%%%%%%%%%%%%%%%%%%%%%%%%%%%%%%%%%%%%%%%%%%%%%%%%%%%%%%%
\section{Related Work}
\label{sec:related}

%===============================
\subsection{Traditional TSC Methods}
\label{subsec:traditional}
Traditional methods heavily rely on hand-crafted traffic signal plans or rules. FixedTime~\cite{fixedtime} and GreenWave~\cite{greenwave} control necessitate predetermined cycle lengths, fixed phase sequences, and specific phase splits. Actuated control, such as self-organizing traffic lights~\cite{sotl2013}, activates traffic signals based on predefined rules and real-time traffic data. Adaptive control, showcased in TSC systems like SCATS~\cite{scats} and SCOOT~\cite{scoot}, involves formulating a set of traffic plans and selecting the optimal one for the current traffic situation based on data from loop sensors. Optimization-based control methods, such as Max Pressure~\cite{mp2013}, conceptualize TSC as an optimization problem within a specific traffic model, adjusting the traffic signal based on observed traffic data. Recently, certain optimization-based approaches, including Max Pressure and Max-QueueLength~\cite{ql}, have supported multi-intersection TSC, demonstrating superior performance compared to RL methods. Advanced-MaxPressure~\cite{advanced}, merging principles from actuated control and optimization-based strategies, has attained a SOTA position in the traditional TSC field, surpassing the performance of the majority of RL methods.

%===============================
\subsection{Online RL-based TSC Methods}
\label{subsec:online}
Online RL approaches are commonly employed in TSC. Typically, various methods develop network structures to enhance TSC performance. FRAP~\citet{frap} designs a specific network structure to construct phase features and capture phase competition relations with expert manual design, making it capable of handling unbalanced traffic flows. AttetnionLight~\cite{ql} utilizes attention mechanisms to automatically model phase correlation. CoLight~\cite{colight} adapts a graph attention network~\cite{gats} for intersection cooperation. AttendLight~\cite{attend} employs the attention mechanism to handle various topologies of intersections.

Furthermore, certain methods emphasize the development of effective state representations or reward function to elevate performance. PressLight~\cite{presslight} undergoes enhancements in LIT~\cite{LIT} and IntelliLight~\cite{intellilight} by integrating “pressure" into the state and reward function design. AttentionLight~\cite{ql} utilizes the queue length as both state representation and reward function, leading to an improvement over FRAP. 
Advanced-XLight~\cite{advanced} utilizes the effective running vehicle number as a state representation. 

Other RL techniques can also contribute to improving TSC performance. DemoLight~\cite{demolight} utilizes imitation learning~\cite{imitation} to expedite the learning process. HiLight~\cite{hilight} empowers each agent to learn a high-level policy, optimizing the objective locally with hierarchical RL~\cite{hierarchical}. MetaLight~\cite{metalight} employs Meta-learning~\cite{meta} to adapt quickly and stably to new traffic scenarios. 

%===============================
\subsection{Offline RL Algorithms}
\label{subsec:offline_rl}
A central concern in offline RL is the extrapolation error, which is the challenge of accurately assessing actions beyond the training distribution. Batch-constrained deep Q-learning (BCQ)~\cite{BCQ} constrains the action space to align closely with on-policy data distributions, aiming to reduce extrapolation errors in fixed batch data scenarios. Behavior regularized actor-critic~\cite{BRAC} mitigates data distribution issues by regulating the policy to closely mirror the behavioral patterns observed in the dataset. Bootstrapping error accumulation reduction~\cite{BEAR} tackles bootstrapping errors by constraining action selection. This approach ensures robust learning from fixed data, overcoming limitations present in Q-learning and actor-critic methods, especially in data-sensitive scenarios. CQL~\cite{CQL} addresses issues of overestimation and distributional shifts by acquiring a conservative Q-function, which improves policy performance through the application of a straightforward Q-value regularizer designed for complex data distributions.

Behavior cloning (BC) offers a direct imitation learning approach, allowing an agent to replicate actions demonstrated by expert. This simplifies the learning process by directly mapping states to actions, avoiding the complexities associated with intricate RL strategies. TD3+BC~\cite{TD3BC} introduces minimal modifications by incorporating a BC term in the policy update and implementing state input normalization, specifically addressing value estimation errors. Both Trajectory Transformer (TT)~\cite{TT} and Decision Transformer (DT)~\cite{DT} reinterpret offline RL as a sequence modeling challenge, harnessing the power of the transformer architecture. TT employs beam search to model trajectory distributions across various RL domains, while DT streamlines RL by generating optimal actions through a causally masked Transformer, conditioned on desired returns, past states, and actions. However, limited offline RL algorithms have been considered into TSC. 

This study introduces DataLight, an innovative offline data-driven approach meticulously crafted to boost the efficiency of TSC. Through the strategic utilization of effective state representations and reward function, DataLight endeavors to optimize traffic signal operations. This optimization is anticipated to result in improved traffic flow, decreased congestion, and an ultimate enhancement in the overall performance of transportation systems.

%%%%%%%%%%%%%%%%%%%%%%%%%%%%%%%%%%%%%%%%%%%%%%%%%%%%%%%%%%%%%%%%%%%%%%%%%%%%%%%%%%
\section{Preliminary}
\label{sec:preliminary}
\begin{figure}[ht]
\centering
\includegraphics[width=1\linewidth]{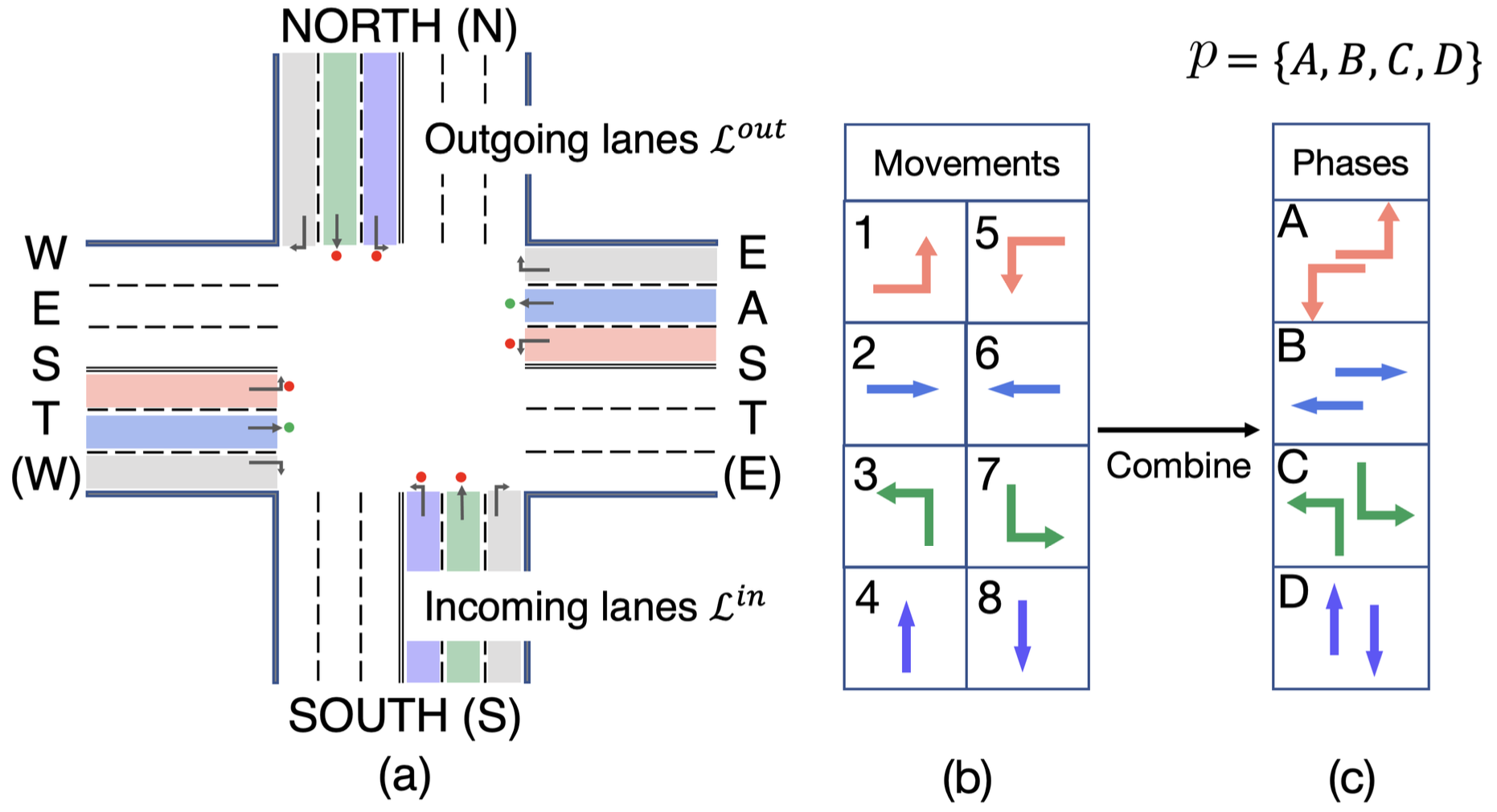}
\caption{Illustration of intersection structure and traffic signal phases: (a) A standard four-way, three-lane intersection. (b) Eight distinct traffic movements. (c) Four signal phases.}
\label{fig:inter}
\end{figure}

%===============================
\paragraph{Traffic signal control.}
TSC involves a comprehensive understanding the traffic environment, encompassing elements such as the traffic network, vehicle movements, signal phases, and state representations of traffic. Typically, a traffic network is depicted as a directed graph, with intersections and roads corresponding to nodes and edges, respectively. Roads consist of three lanes, serving as the primary channels for vehicle movement. State representations are lane-based to provide a more accurate description of current states. 

Formally, the set of incoming and outgoing lanes at intersection $i$ can be represented as $\mathcal{L}_i^{in}$ and $\mathcal{L}_i^{out}$, respectively. Vehicle movements are categorized by their directional paths: left, straight, and right. Each signal phase $p$ is a combination of these traffic movements that are simultaneously permitted. The set of all phases is denoted by $\mathcal{P}$. For each phase $p$, $\mathcal{L}_{p}^{in}$ indicates the involved incoming lanes. As illustrated in Figure~\ref{fig:inter}, in a standard four-way, three-lane intersection, there are typically four signal phases, each with distinct $\mathcal{L}_{p}^{in}$. The regulation of vehicle passage, achieved by managing the signal lights for each phase at each intersection, allowed for the effective control of the entire traffic system.

%===============================
\paragraph{Offline RL.}
Consider learning within a Markov decision process (MDP) described by the tuple $(\mathcal{S}, \mathcal{A}, P, \mathcal{R}, \gamma)$, where states $s \in \mathcal{S}$, actions $a \in \mathcal{A}$, transition dynamics $P(s|s, a)$, a reward function $r\in \mathcal{R}$, and a discount factor $\gamma$. Let $s_t$, $a_t$, and $r_t$ represent the state, action, and reward at timestep $t$. The goal is to formulate a policy that maximizes the expected return $\mathbb{E}\left[\sum_{t=1}^T r_t\right]$ within the MDP. Offline RL involves learning a policy from a predetermined dataset, which includes single-step transitions $\mathcal{D}={(s_t, a_t, s_{t+1}, r_t)}$ without active interaction with the environment. This setup restricts agents from exploring the environment and acquiring additional feedback, introducing complexity to the learning process.

%===============================
\paragraph{Problem definition.} 
Three distinct TSC methods interacting with the environment result in three offline datasets with distinct quality. These datasets capture the historical trajectories represented as $(s_t, a_t, s_{t+1}, r_t)$ at each timestep $t$. The RL agent learns an optimal policy to maximize the expected reward across historical trajectories, denoted by
\begin{equation}
\max _{\hat{\pi}} \mathbb{E}_{(s \sim \mathcal{D},a=\hat{\pi}(s))}[R(s, a)],
\end{equation}
where $\hat{\pi}(\cdot)$ represents the final learned behavioral policy, denoting the argmax operation in this study. The ultimate goal is to apply the optimally learned policy, derived from offline RL, in scenarios involving multiple intersections. A single agent manages all intersections, ensuring efficient and coordinated traffic control. This approach integrates learning from diverse datasets to optimize signal control in complex, real-world traffic systems.

%%%%%%%%%%%%%%%%%%%%%%%%%%%%%%%%%%%%%%%%%%%%%%%%%%%%%%%%%%%%%%%%%%%%%%%%%%%%%%%%%%
\section{DataLight}
\label{sec:dataLight}
%===============================
\subsection{State and Reward Design}
\label{subsec:state_reward_design}
%===============================
\paragraph{State representations.}
Vehicle speed serves as a crucial indicator of traffic flow dynamics. Specifically, let $V(x)/V_{max}$ denotes the velocity saturation degree, while $1-V(x)/V_{max}$ indicates the unsaturated degree of each vehicle $x$. Here, $V_{max}$ represents the maximum permitted velocity. These metrics characterize both static and dynamic states of vehicles. A saturation degree of 1 denotes a vehicle at full speed, while 0 suggests that the vehicle is stopped.

To better capture traffic dynamics, roads near the intersection, extending for 400 meters, are divided into eight 50-meter segments. The vehicle count in each section is monitored, and this segmentation is crucial for model performance, as detailed in Appendix~\ref{apxsec:segs}. Additionally, we compute the aggregate of both the velocity saturation and unsaturation degrees across these segments to create  comprehensive state representations, including the number of vehicles, the sum of velocity saturation degrees, and the sum of unsaturation degrees under each segmented road section. Note that the total number of vehicles corresponds to the combined sum of the saturation and unsaturation degrees in these segments. Finally, the state representations include the current phase, the number of vehicles, and the total velocity saturation and unsaturation degrees under each road segment. These states provide a detailed and dynamic view of traffic conditions, essential for accurate traffic modeling and the implementation of effective control strategies.

%===============================
\paragraph{Reward function.}
Within a 200-meter proximity of the intersection, the reward is calculated as the sum of velocity unsaturation degrees for each approaching lane.
\begin{equation}
r = -\sum_x \left (1-\frac{V(x)}{V_{max}} \right ),
\end{equation}
where $x$ represents a vehicle within 200-meter range, $V(x)$ denotes its current velocity, and $V_{max}$ is the maximum permitted velocity. 

Data from lanes allocated for right turns are excluded from this calculation, recognizing that velocity reductions in these lanes are typically not influenced by the TSC system. The limitation of the assessment range to 200 meters ensures that vehicle deceleration caused by controls at adjacent intersections does not impact the analysis. A higher reward value indicates a reduction in the number of vehicles affected (either stopped or slowed down) by the intersection signal. This solution effectively captures both the stationary queueing vehicles at intersections and the moving vehicles approaching intersections. It outlines the impact of intersections on traffic flow, considering both stationary and moving vehicles within a strategically determined range.

%===============================
\paragraph{Action space.}
Each action corresponds to actuating one phase, allowing vehicles for the respective traffic movements to pass through the intersection. The action space consists of four phases, denoted as $\{A, B, C, D\}$, as illustrated in Figure \ref{fig:inter} (c). The RL agent decision-making process involves selecting one of these actions during each fixed action duration at time step $t$. The RL agent decides to maintain the current phase if $a_t = a_{t-1}$, and switch to anther one for the next action duration if $a_t\neq a_{t-1}$.

%===============================
\subsection{Network Structure Design}
\label{subsec:network}
%===============================
\paragraph{Spatial encoding (SE).}
The state inputs undergo an initial embedding layer and are then concatenated based on their spatial correlation, as follows:
\begin{align}
\mathbf{F}_{l,i} = \operatorname{Emb}(s_{l, i})\oplus \operatorname{Emb}(s_{l, i+8})\oplus \operatorname{Emb}(s_{l, i+16}),    
\end{align}
where $i$ ranges from $0$ to $7$, corresponding to the 8 segments, and $\oplus$ denotes the concatenate operation. Then, multi-head self-attention (MHA) is employed for SE.
\begin{align}
\mathbf{F}_l = \operatorname{MHA}\left (\mathbf{F}_{l,0} \oplus \mathbf{F}_{l,1} \oplus \cdots \oplus\mathbf{F}_{l,7} \right).
\end{align}

%===============================
\paragraph{Feature fusion.}
The lane features are modeled using MHA based on the phase composition, followed by averaging to obtain the phase feature. The phase feature is then subjected to another MHA operation to capture phase correlations. The resulting correlated phase feature is embedded to derive the Q-values. Additional details of the network design are provided in Appendix~\ref{apxsec:network}.

%===============================
\subsection{Learning Procedure}
\label{subsec:model_learning}
DataLight undergoes an update process that combines losses of temporal difference (TD), eigensubspace regularization (ER)~\cite{ER}, and CQL~\cite{CQL}.

%===============================
\paragraph{TD loss.} The TD loss, also known as the Bellman error, is calculated as follows:
\begin{align}
\mathcal{L}_{TD}(\bm{\theta})  = \mathbb{E}_{(s, a, r, s^{\prime}) \sim \mathcal{D}}\left[y-Q(s, a;\bm{\theta})\right]
\end{align}
where $\bm{\theta}$ and $Q(\cdot)$ represent the parameters of the deep Q-learning neural network and its Q-value function, respectively. $y$ denotes the target value, which is calculated by
\begin{align}
y = r+\gamma \operatorname{max}_{a^\prime}Q(s^\prime, a^\prime; \bm{\theta}^-),
\end{align}
where $s^\prime$ and $a^\prime$ denote the next state and action, respectively. $\gamma$ represents the discount factor, and $\theta^-$ is the parameters of the target neural network.

%===============================
\paragraph{ER loss.}
At each time step, the ER loss is adjusted towards the 1-eigensubspace, and the regularization term is minimized as follows:
\begin{align}
\mathcal{L}_{ER}(\bm{\theta}) = \frac{1}{N}\sum_{n=1}^{N}\left\|B_n(\bm{\theta})-Z\right\|_2^2,
\end{align}
where $B_n(\cdot)$ represents the Bellman error at the $n$-th dimension, and $Z = \frac{1}{N}\sum_{n=1}^{N}B_n(\bm{\theta})$. $\left\|\cdot \right\|_2^2$ denotes the squared Euclidean norm calculation.

%===============================
\paragraph{CQL loss.} 
CQL loss is calculated by
\begin{equation}
\begin{split}
\mathcal{L}_{CQL}(\bm{\theta}) &= \min _Q  \mathbb{E}_{s \sim \mathcal{D}}\biggl[\log \sum_{a} \exp \left(Q(s,a)\right) \biggr.\\
&\quad -\biggl. \mathbb{E}_{a \sim \pi(a \mid s)}\left[Q(s, a)\right]\biggr].
\end{split}
\end{equation}

Finally, the total loss of DataLight is formulated as a combination of TD, ER, and CQL losses, with the balance regulated by hyperparameters $\alpha$ and $\beta$:
\begin{align}
\mathcal{L}(\bm{\theta}) = \mathcal{L}_{TD}(\bm{\theta})+\alpha \mathcal{L}_{ER}(\bm{\theta})+\beta\mathcal{L}_{CQL}(\bm{\theta}).
\end{align}

%%%%%%%%%%%%%%%%%%%%%%%%%%%%%%%%%%%%%%%%%%%%%%%%%%%%%%%%%%%%%%%%%%%%%%%%%%
\section{Experiments}
\label{sec:exp}
We conducted experiments on CityFlow~\cite{cityflow} to answer the following research questions (RQ):
\begin{itemize}[leftmargin=*]
\item \textbf{RQ1}: How does the performance of DataLight compare with that of SOTA approaches?
\item \textbf{RQ2}: How do the techniques employed in DataLight, including state and reward design, and SE, impact the learning process?
\item \textbf{RQ3}: How scalable is DataLight in addressing real-world application issues?
\end{itemize}

%===============================
\paragraph{Datasets.}
Three real-world traffic flow datasets from JiNan (JN), HangZhou (HZ), and New York (NY) were utilized, comprising seven distinct datasets in total (i.e., JN1, JN2, JN3, HZ1, HZ2, NY1, and NY2). We evaluated the performance of DataLight on these datasets, with each dataset offering distinct intersection layouts and traffic flows. JN, HZ, and NY datasets featured a 12-, 16-, and 192-intersection grid. With varying traffic patterns and intersection configurations, these datasets provided a comprehensive test bench to ensure a thorough evaluation of DataLight. The following approaches were employed to generate offline datasets from the JN and HZ datasets:
\begin{itemize}[leftmargin=*]
\item \textbf{Random offline data (ROD):} A random policy is used to generate a low-quality offline dataset.
\item \textbf{Medium offline data (MOD):} FRAP is employed to generate a medium-quality offline dataset, reflecting its intermediate performance level.
\item \textbf{Expert offline data (EOD):} Advanced-CoLight is utilized to produce a high-quality expert-level dataset. 
\item \textbf{Cyclical offline data (COD):} FixedTime is utilized to create an offline dataset with a specific structure, where actions follow a cyclical pattern. This structured data, reflecting real-world scenarios, is utilized to address practical application issues associated with offline methods.
\end{itemize}

These offline datasets were created using methods that vary in performance, resulting in datasets of different quality. For additional details, refer to Appendix~\ref{apxsec:data}. For the ROD, MOD, and EOD offline datasets, 10 episodes were conducted for each of the five traffic flow datasets, extracting 20K tuples from each (totaling approximately 100K tuples) and discarding excess data. For the COD, one episode generated 11240 transitions at a 20-second action duration. 

Furthermore, the offline datasets can be generated at different action durations (refer to Appendix~\ref{apxsec:data}), and DataLight exhibits different performance on these datasets (see Appendix~\ref{apxsec:model_performance}). The performance is evaluated under JN and HZ datasets using an offline dataset with an action duration of 10 seconds, while for NY, 20 seconds are used.

%===============================
\paragraph{Hyperparameters.}
The Adam optimizer was utilized with learning rates of $1\mathrm{e}{-3}$ and $2\mathrm{e}{-4}$, respectively. The discount factor $\gamma$ was set to 0.8. The training process comprises 120 episodes, each undergoing two training epochs. To monitor the training progress, model evaluations were conducted at end of each episode. For a fair comparison, hyperparameters for all offline and online RL baselines were set identical to those of DataLight and AttentionLight, respectively.

%===============================
\begin{table*}[ht]
\setlength\tabcolsep{1.5pt}
\centering
\begin{threeparttable}
\caption{Comparison results with traditional, online RL, and offline RL-based baseline models. A smaller ATT (in seconds) indicates better overall performance evaluation.}
\label{tab:overall}
\begin{tabular}{clc|ccc|cc|cc}\toprule
\multirow{2}{*}{} & \multicolumn{2}{l}{\multirow{2}{*}{Model}} & \multicolumn{3}{c}{JN} & \multicolumn{2}{c}{HZ} & \multicolumn{2}{c}{NY} \\\cmidrule{4-10} 
& & & JN1 & JN2 & JN3 & HZ1 & HZ2 & NY1 & NY2 \\\midrule

\multirow{3}{*}{Traditional} & \multicolumn{2}{l|}{FixedTime~\cite{fixedtime}} & 429.27 & 370.34 & 384.89 & 497.87 & 408.31 & 1507.12 & 1733.30 \\
& \multicolumn{2}{l|}{MaxPressure~\cite{mp2013}} & 269.87 & 239.75 & 240.03 & 284.44 & 327.62 & 1122.00 & 1462.96 \\
& \multicolumn{2}{l|}{Max-QueueLength~\cite{ql}} & 268.87 & 240.02 & 238.51 & 284.32 & 325.44 & 1197.59 & 1551.46 \\\midrule

\multirow{6}{*}{\makecell{Online\\RL-\\based}} & \multicolumn{2}{l|}{FRAP~\cite{frap}} & 299.06 & 266.19 & 273.58 & 321.88 & 353.89 & 1192.23 & 1470.51 \\
& \multicolumn{2}{l|}{MPLight~\cite{mplight}} & 297.68 &	274.32 & 268.00 & 313.16 & 355.35 &	1321.40 & 1642.05 \\
& \multicolumn{2}{l|}{PRGLight~\cite{prglight}} & 291.27 & 257.52 & 261.74 & 301.06 & 369.98 & 1283.37 & 1472.73 \\
& \multicolumn{2}{l|}{CoLight~\cite{colight}} & 269.47 & 252.35 & 249.05 & 297.64 & 337.25 & 1065.64 & 1367.54 \\
& \multicolumn{2}{l|}{AttentionLight~\cite{ql}} & 256.36 & 240.10 & 237.00 & 284.85 & 313.89 & 1013.78 & 1401.32 \\
& \multicolumn{2}{l|}{Advanced-CoLight~\cite{advanced}} & 246.42 & 233.68 & 229.61 & 271.62 & 311.07 & 970.05 & 1300.62 \\\midrule 
    
\multirow{12}{*}[-8pt]{\makecell{Offline\\RL-\\based}} & \multirow{3}{*}{BC~\cite{TD3BC}} & ROD & 1488.54 & 1620.43 & 1466.24 & 1635.70 & 1091.74 & 2219.42 & 2354.75 \\
& & MOD & 439.87 & 406.43 &	398.07& 522.55 & 454.22 & 1273.95 &	1591.58 \\
& & EOD &246.89 & 225.37 & 223.81 & 263.80 & 319.91 & 1184.07 &	1480.33 \\\cmidrule{2-10} 

& \multirow{3}{*}{BCQ~\cite{BCQ}} & ROD & 267.02 & 237.72 &	237.24 & 287.18 & 338.54 & 1151.87 & 1471.87 \\
& & MOD & 269.50 & 233.84 & 237.97 & 274.84 & 338.15 & 1045.30 & 1362.12 \\
& & EOD & 253.25 & 229.51 & 229.16 & 267.30 & 317.99 & 1061.24 & 1398.21 \\\cmidrule{2-10} 

& \multirow{3}{*}{CQL~\cite{CQL}} & ROD & 275.12 & 240.87 &	242.77 & 285.28 & 347.48 & 1195.83 & 1528.99 \\
& & MOD & 251.82 & 234.31 &	231.73 & 271.20 & 326.05 & 1115.47 & 1436.03 \\
& & EOD & 252.37 & 236.83 & 234.33 & 280.84 & 317.82 & 1080.97 & 1421.87 \\\cmidrule{2-10} 
    
& \multirow{3}{*}{\textbf{DataLight}} & ROD & $\colorbox[gray]{0.9}{238.68}$ & $\colorbox[gray]{0.9}{223.88}$ & $\colorbox[gray]{0.9}{221.36}$ & $\colorbox[gray]{0.9}{263.24 }$ & $\colorbox[gray]{0.9}{301.40}$ & 1095.40 	&1367.04 \\
& & MOD & $\colorbox[gray]{0.9}{239.13}$ &	$\colorbox[gray]{0.9}{225.16}$ & $\colorbox[gray]{0.9}{222.60}$ & 	$\colorbox[gray]{0.9}{264.75}$ &	$\colorbox[gray]{0.9}{297.75}$ &	$\colorbox[gray]{0.9}{920.04}$ & $\colorbox[gray]{0.9}{1221.33}$ \\
& & EOD & $\colorbox[gray]{0.9}{ 234.94}$ & $\colorbox[gray]{0.9}{221.31}$ & $\colorbox[gray]{0.9}{218.57}$ &	$\colorbox[gray]{0.9}{261.56}$ &	$\colorbox[gray]{0.9}{298.18}$ &	1075.23 & 1393.08 \\\bottomrule
\end{tabular}
\begin{tablenotes}
\footnotesize
\item[$\star$] The values in the gray cells represent the maximum scores among all the different models.
\end{tablenotes}
\end{threeparttable}
\end{table*}

%===============================
\paragraph{Evaluation metric.}
The average travel time (ATT)~\cite{survey}, a commonly used metric for TSC, is employed to evaluate the performance of DataLight. We performed all experiments three times and calculated the average ATT for the last 10 and 5 episodes across the three repetitions for the online and offline methods, respectively.

%===============================
\subsection{Overall Performance Evaluation Results (RQ1)}
\label{subsec:overall}
We conducted a comparative analysis of DataLight against various baseline models, encompassing traditional approaches (FixedTime, Max Pressure, and Max QueueLength), online RL-based methods (FRAP, MPLight, PRGLight, CoLight, AttentionLight, and Advanced-CoLight), and offline RL-based methods (BC, BCQ, and CQL). Note that the offline RL-based methods leverage the state representations and reward function from Advanced-XLight and employ the network structure of AttentionLight. Additionally, both DT~\cite{DT} and TT~\cite{TT} are excluded as offline baselines due to their inability to properly perform the TSC tasks (see Appendix~\ref{apxsec:DT}).

Table~\ref{tab:overall} demonstrates the performance comparison of ATT on real-world datasets. DataLight demonstrates superior performance compared to all traditional and online RL baselines when trained on JN, HZ, and NY datasets. Specifically, on the JN1, JN2, and JN3 datasets, DataLight achieves significant improvements in ATT metric by 4.9\%, 5.6\%, and 5.1\%, surpassing the SOTA Advanced-CoLight. Similarly, DataLight improved by 3.8\%, 4.5\%, 5.4\%, and 6.5\% on the HZ1, HZ2, NY1, and NY2 datasets. Furthermore, we trained DataLight on the three distinct offline datasets (ROD, MOD, and EOD) and subsequently evaluated its performance on the JN, HZ, and NY datasets. Even when trained on ROD, DataLight significantly outperformed CQL under EOD, achieving improvements exceeding 5.8\% and 5.4\% on the JN and HZ datasets, respectively. For the NY1 and NY2 datasets under MOD, DataLight surpassed BCQ by 13.6\% and 11.5\%, respectively. Overall, DataLight exhibits remarkable potential to surpass SOTA baselines, both in online and offline scenarios, as a pioneering benchmark for addressing challenges in TSC with RL.

%===============================
\subsection{Ablation studies (RQ2)}
\label{subsec:ablation}
We conducted separate evaluations of the effects of state representations, reward function, and SE in DataLight, aiming to individually assess their impact and efficacy on the overall performance.

%===============================
\begin{table}[t]
\centering
\caption{The impact of the designed state representations on performance (ATT in seconds) of DataLight.}
\label{tab:states}
\begin{tabular}{lcc|c}\toprule
\multirow{2}{*}{Model} & \multicolumn{2}{c}{JN} & \multicolumn{1}{c}{HZ} \\\cmidrule {2-4} 
&JN1&JN2&HZ1\\\midrule
w/o designed states & 244.32 & 226.82 & 266.91 \\
\textbf{DataLight (ROD)} & $\colorbox[gray]{0.9}{238.68}$ & $\colorbox[gray]{0.9}{223.88 }$ & $\colorbox[gray]{0.9}{263.24}$	\\
Improvement & $\uparrow$ 2.31\% & $\uparrow$ 1.30\% & $\uparrow$ 1.37\% \\\midrule
w/o designed states & 244.12 &	228.50 & 269.55 \\
\textbf{DataLight (MOD)} & $\colorbox[gray]{0.9}{239.13}$ & $\colorbox[gray]{0.9}{225.16}$ &	$\colorbox[gray]{0.9}{264.75}$ \\
Improvement & $\uparrow$ 2.04\% & $\uparrow$ 1.46\% & $\uparrow$ 1.78\% \\\midrule
w/o designed states & 239.75 & 224.76 & 264.03  \\
\textbf{DataLight (EOD)} & $\colorbox[gray]{0.9}{234.94}$ & $\colorbox[gray]{0.9}{221.31}$ &	$\colorbox[gray]{0.9}{261.56}$ \\
Improvement & $\uparrow$ 2.01\% & $\uparrow$ 1.53\% & $\uparrow$ 0.94\% \\\bottomrule
\end{tabular}
\end{table}

\paragraph{Effects of the designed state representations.}
The effectiveness of the designed state representations is illustrated in Table~\ref{tab:states} and Table~\ref{apxtab:states}. Note that “w/o designed states" denotes that the state representations include only the current phase, the number of vehicles on entering and exiting lanes, queue length on entering and exiting lanes, traffic movement pressure, running vehicles count, and queueing vehicle count. Compared to DataLight without the designed state representations, DataLight with the state representations improved ATT by 2.31\%, 1.30\%, and 1.37\% on the JN1, JN2, and HZ1 datasets when trained on MOD. It also exhibited improvements of 2.04\%, 1.46\%, and 1.78\% on the JN1, JN2, and HZ1 datasets when trained on MOD, and improvements of 2.01\%, 1.53\%, and 0.94\% on EOD. The comparative results indicate the effectiveness of the designed innovative state representations of DataLight, which can significantly enhance the performance of offline data-driven RL strategies.

%===============================
\begin{table}[t]
\centering
\caption{The impact of the designed reward function on performance (ATT in seconds) of DataLight.}
\label{tab:reward}
\begin{tabular}{lcc|c}\toprule
\multirow{2}{*}{Model} & \multicolumn{2}{c}{JN} & \multicolumn{1}{c}{HZ} \\\cmidrule {2-4} 
& JN1 & JN2 & HZ2 \\\midrule
w/ pressure & 369.63 & 373.04 & 355.23  \\
w/ queue length &251.48 & 230.65 & 312.07 \\
\textbf{DataLight (ROD)} &$ \colorbox[gray]{0.9}{238.68}$ &	$\colorbox[gray]{0.9}{223.88}$ & $\colorbox[gray]{0.9}{301.40} $ \\
Improvement & $\uparrow$ 5.09\% & $\uparrow$ 2.94\% & $\uparrow$ 3.42\% \\\midrule
w/ pressure & 269.00 & 238.42 &	318.31 \\
w/ queue length & 243.48 & 225.68 &	305.17 \\
\textbf{DataLight (MOD)} & $\colorbox[gray]{0.9}{239.13}$ &	$\colorbox[gray]{0.9}{225.16}$ & $\colorbox[gray]{0.9}{297.75}$ \\
Improvement & $\uparrow$ 1.79\% & $\uparrow$ 0.23\% & $\uparrow$ 2.43\% \\\midrule
w/ pressure & 259.16 & 230.19& 311.20 \\
w/ queue length & 238.24 & 222.70 &	299.15 \\
\textbf{DataLight (EOD)} & $\colorbox[gray]{0.9}{234.94}$ &	$\colorbox[gray]{0.9}{221.31}$ & $\colorbox[gray]{0.9}{298.18}$ \\
Improvement & $\uparrow$ 1.39\% & $\uparrow$ 0.62\% & $\uparrow$ 0.32\% \\\bottomrule
\end{tabular}
\end{table}

\paragraph{Effects of the designed reward function.}
To delve into the performance of DataLight, we scrutinized its correlation with the designed reward function. The effects are showcased in Tables~\ref{tab:reward} and~\ref{apxtab:reward}. Compared to DataLight without the designed reward function, the version with the designed reward function exhibited notable improvements in ATT: 5.09\%, 2.94\%, and 3.42\% on JN1, JN2, and HZ2 when trained on ROD, 1.79\%, 0.23\%, and 2.43\% when trained on MOD, and 1.39\%, 0.62\%, and 0.32\% when trained on EOD. Moreover, particularly when trained on ROD, the designed reward function significantly enhanced DataLight's performance compared to improvements observed on both MOD and EOD. These results highlight that DataLight with the designed reward function can maintain comparable effectiveness on EOD even when trained on ROD, a crucial factor for achieving a generalized model trained on diverse offline datasets.

%===============================
\begin{table}[t]
\centering
\caption{The impact of SE on performance (ATT in seconds) of DataLight.}
\label{tab:STE}
\begin{tabular}{lcc|c}\toprule
\multirow{2}{*}{Model} & \multicolumn{2}{c}{JN} & \multicolumn{1}{c}{HZ} \\\cmidrule {2-4} 
& JN1 & JN2 & HZ1 \\\midrule
w/o SE & 242.15 & 225.53 & 264.26 \\
\textbf{DataLight (ROD)} & $\colorbox[gray]{0.9}{238.68}$ &	$\colorbox[gray]{0.9}{223.88}$ & $\colorbox[gray]{0.9}{263.24}$ \\
Improvement & $\uparrow$ 1.43\% & $\uparrow$ 0.72\% & $\uparrow$ 0.39\% \\\midrule
w/o SE & 240.02 & 225.16 &	$\colorbox[gray]{0.9}{264.69}$ \\
\textbf{DataLight (MOD)} & $\colorbox[gray]{0.9}{239.13}$ & 225.16 & 264.75 \\
Improvement & $\uparrow$ 0.37\% & $\uparrow$ 0.00\% & $\downarrow$ 0.02\% \\\midrule
w/o SE & 236.51 & 221.93 &	262.29 \\
\textbf{DataLight(EOD)} & $\colorbox[gray]{0.9}{234.94}$ &	$\colorbox[gray]{0.9}{221.31}$ & $\colorbox[gray]{0.9}{261.56}$ \\
Improvement & $\uparrow$ 0.66\% & $\uparrow$ 0.28\% & $\uparrow$ 0.28\% \\\bottomrule
\end{tabular}
\end{table}

\paragraph{Effects of spatial encoding.}
We also assessed the impact of SE on the performance of DataLight, detailed in Table~\ref{tab:STE} and Table~\ref{apxtab:STE}. In comparison to the model without SE, DataLight with SE showcased improvements in ATT by 1.43\%, 0.72\%, and 0.39\% on JN1, JN2, and HZ1 when trained on ROD, by 0.37\% on JN1 when trained on MOD, and by 0.66\%, 0.28\%, and 0.28\% on JN1, JN2, and HZ1 when trained on EOD. Furthermore, with the incorporation of SE, DataLight exhibited superior performance on ROD compared to both MOD and EOD. These results affirm the efficacy of SE in enhancing performance.

%===============================
\subsection{Case studies (RQ3)}
\label{subsec:case}
%===============================
\begin{figure}[t]
\centering
\includegraphics[width=1\linewidth]{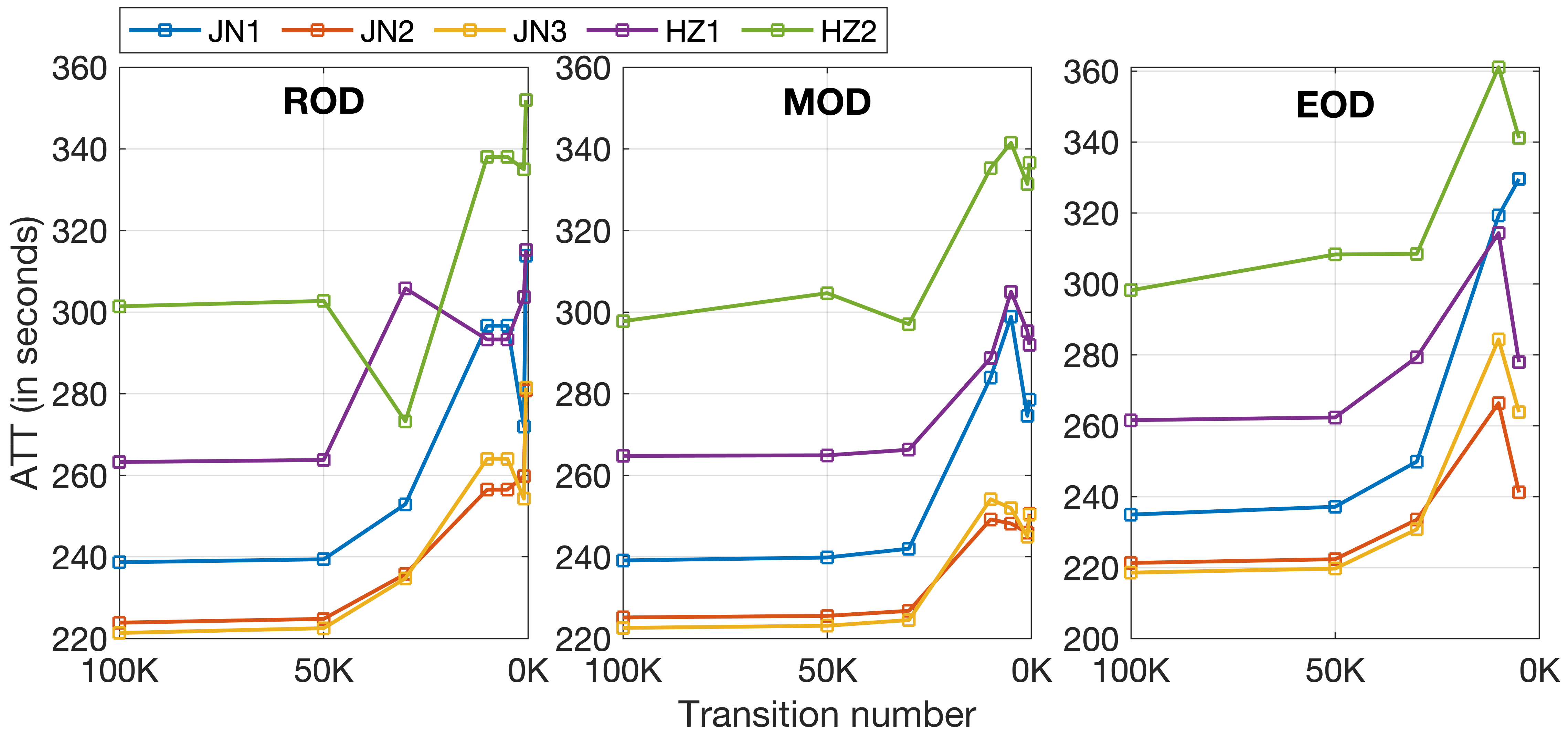}
\caption{Performance evaluation of DataLight with varying amounts of offline data (ATT in seconds).}
\label{fig:lowdata}
\end{figure}

\paragraph{Low-data scenario.}
We evaluated the learning capabilities of DataLight under varying amounts of offline datasets, ranging from 100K, 50K, 30K, 10K, 5K, 1K, to 0.5K. Figure~\ref{fig:lowdata} illustrates the impact of data volume on the performance of DataLight. Notably, when the data volume exceeded 30K, the ATT of DataLight trained with the data only showed a slight diminishing effect. However, a noticeable reduction in model effectiveness occurs when the dataset size is less than 10K. Despite this, DataLight maintains satisfactory performance, exhibiting only a minimal decrease compared to the optimal results. Note that when trained on EOD with less than 5K data, DataLight failed to achieve stable results. This suggests that a larger dataset is more suitable for stabilizing the training on EOD. These findings underscore the reliability of DataLight, showcasing its robust capability to learn well even with limited data.

%===============================
\begin{figure}[t]
\centering
\includegraphics[width=1\linewidth]{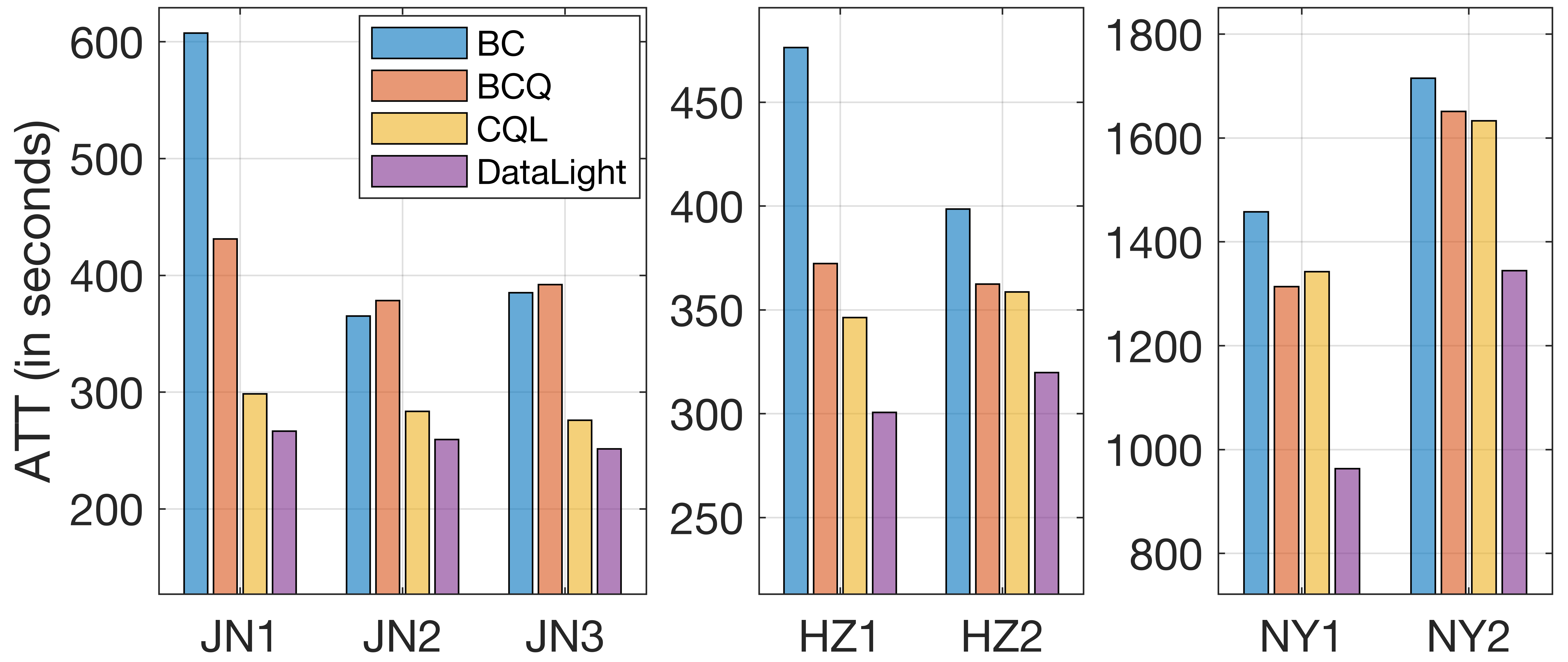}
\caption{Performance of DataLight on COD (ATT in seconds).}
\label{fig:COD}
\end{figure}

\paragraph{Learning scenario from COD.}
In real-world TSC scenarios, FixedTime stands out as the most commonly employed method. To tackle the challenges of real-world learning, we gathered transitions under FixedTime control, each with an action duration of 15 seconds from JN and HZ datasets, creating COD offline data. Figure~\ref{fig:COD} and Table~\ref{apxtab:COD} illustrate the performance of DataLight when trained on COD. The ATT results of DataLight (depicted in purple) were below 300, 350, and 1400 seconds for the JN, HZ, and NY datasets. Furthermore, DataLight demonstrated a substantial performance advantage over all other offline RL models (BC, BCQ, and CQL), emphasizing its robust learning capabilities from COD. Overall, these results strongly indicate the potential efficacy of deploying and applying DataLight in real-world scenarios.

To thoroughly assess the efficacy of DataLight, we conducted an evaluation that involved the collection and amalgamation of various COD. Subsequently, DataLight was trained on this aggregated dataset and subjected to testing across a range of action durations. The results of this comprehensive evaluation are presented in Table~\ref{tab:CODm} and Table~\ref{apxtab:CODm}, providing detailed insights into the model's performance metrics. The outcomes highlight DataLight's remarkable level of performance, underscoring its robust learning capabilities. Moreover, in specific scenarios, DataLight surpassed the performance of the Advanced-CoLight, further emphasizing its superior performance characteristics.

\begin{table}[t]
\setlength\tabcolsep{12pt}
\centering
\caption{Performance evaluation of DataLight under varying action durations when trained with mixed COD (ATT in seconds).}
\label{tab:CODm}
\begin{tabular}{cc|c|c}\toprule
Duration & JN1 & HZ1 & NY1 \\\midrule 
10 & 264.74 & $\colorbox[gray]{0.9}{266.25}$ & 978.94 \\
15 & $\colorbox[gray]{0.9}{250.19}$ & 271.46 & 959.41 \\
20 & 253.12 & 281.09 & 969.19 \\
25 & 263.65 & 292.70 & $\colorbox[gray]{0.9}{922.30}$ \\\bottomrule
\end{tabular}
\end{table}

Additionally, we explored the impact of various action durations on the performance of DataLight. Specifically, we collected and amalgamated various COD. Subsequently, DataLight was trained on this aggregated dataset and subjected to testing across a range of action durations (i.e., 10, 15, 20, and 25 seconds). The results are presented in Table~\ref{tab:CODm} and Table~\ref{apxtab:CODm}. The outcomes reveal that DataLight achieved a notably high level of performance, thereby underscoring its robust learning capabilities. Furthermore, in certain scenarios, DataLight can match and exceed the performance of the Advanced-CoLight model, thereby reinforcing its superior performance characteristics. 
%\rev{In mixed COD environments, DataLight's performance varies with different action durations depending on the dataset's characteristics. This indicates that the effectiveness of DataLight is influenced by how well the action duration aligns with each specific dataset, emphasizing the need for dataset-specific customization.}

%===============================
\section{Conclusion}
\label{sec:conclusion}
This study introduced DataLight, an innovative offline data-driven RL-based approach to TSC. DataLight incorporates effective state representations and a reward function by capturing vehicular speeds and segment road spaces within the TSC environment. The dynamic and spatial representation can enhance traffic conditions through spatial sequential modeling. Experimental results demonstrated that DataLight outperforms SOTA online and offline TSC models. Additionally, DataLight exhibited robust learning capabilities for real-world deployment, evident in two key aspects. First, DataLight achieved satisfactory control results even with limited data. Second, when trained on the easily accessible real-world dataset COD, DataLight significantly outperformed all other offline models.

DataLight has two main limitations. (1) DataLight cannot fully investigate the influence of offline datasets with different action durations on model performance. (2) DataLight lacks the capability to consider the augmentation of neighbor information for policy determination. In future work, we will collect real-world traffic data and develop robust RL models in practical scenarios, establishing a strong connection between academic research and real-world applications.

%%%%%%%%%%%%%%%%%%%%%%%%%%%%%%%%%%%%%%%%%%%%%%%%%%%%%%%%%%%%%%%%%%%%%%%%%%%%%%%
\section*{Impact Statements}
In the context of training models for deployment in real-world scenarios, DataLight holds considerable significance. This importance is attributed to two key factors:
\begin{itemize}[leftmargin=*]
\item Robust Learning Capability: DataLight demonstrates a robust learning capability, effectively acquiring highly efficient strategies from a wide range of offline data of varying quality, including from limited amounts of COD. The capacity to adapt and learn from varied data sources is crucial for the development of robust and versatile models.
\item Effectiveness with Easily Accessible Real-World Data: COD represents data that can be easily collected in the real-world. DataLight showcases commendable learning performance with mixed COD datasets. This versatility indicates that DataLight effectively utilizes data that is both abundant and easily obtainable in real-world settings.
\end{itemize}
These attributes guide us toward a methodology where real-world data can be collected and utilized to train models viable for real-world deployment. DataLight significantly contributes to bridging the gap between academic research and practical applications. It underscores the potential of leveraging real-world data to develop theoretically sound and practically applicable models, enhancing the relevance and impact of academic research in real-world scenarios.

%%%%%%%%%%%%%%%%%%%%%%%%%%%%%%%%%%%%%%%%%%%%%%%%%%%%%%%%%%%%%%%%%%%%%%%%%%%%%%%
\bibliography{reference}
\bibliographystyle{icml2024}

%%%%%%%%%%%%%%%%%%%%%%%%%%%%%%%%%%%%%%%%%%%%%%%%%%%%%%%%%%%%%%%%%%%%%%%%%%%%%%%
% Appendix
%%%%%%%%%%%%%%%%%%%%%%%%%%%%%%%%%%%%%%%%%%%%%%%%%%%%%%%%%%%%%%%%%%%%%%%%%%%%%%%
\clearpage
\onecolumn
\setcounter{secnumdepth}{2}
\numberwithin{figure}{section}
\numberwithin{table}{section}
\numberwithin{equation}{section}

%%%%%%%%%%%%%%%%%%%%%%%%%%%%%%%%%%%%%%%%%%%%%%%%%%%%%%%%%%%%%%%%%%%%%%%%%%%%%%%
\begin{appendices}

\section{Model study}

%===============================
\subsection{Effects of different segments}
\label{apxsec:segs}
DataLight's performance is assessed using different state representations based on different road segmentations. For all offline datasets, actions are set with a duration of 10 seconds. The approach involves segmenting a 400-meter area approaching the intersection into either four or eight segments. The results, as depicted in Table~\ref{apxtab:segs}, highlight DataLight's performance with these diverse state representations. It is observed that DataLight achieves its best performance with the eight segmented states in the JN and HZ datasets. In the NY dataset, the performance differences are minimal. Based on these findings, the eight-segmented state representation is chosen as the default for DataLight.

\begin{table}[ht]
\caption{DataLight performance with different state representations that use four or eight segments (ATT in seconds).}
\label{apxtab:segs}
\centering
\begin{tabular}{lcccccccc}\toprule
\multirow{2}{*}{Segments} &\multirow{2}{*}{Dataset} &\multicolumn{3}{c}{ JN } & \multicolumn{2}{c}{ HZ }  & \multicolumn{2}{c}{ NY } \\
\cmidrule { 3 - 9 } & &JN1 & JN2 & JN3 & HZ1 & HZ2& NY1 & NY2 \\\midrule
\multirow{3}{*}{4} &ROD  & 239.98 &225.61 &	222.86 &	264.51 &	293.08 &	1241.08& 	1571.27  \\
&MOD  & 240.71 &227.48 &	224.67 &	266.49 &	301.21 &	1212.93 	&1569.76 \\
&EOD  & 236.49 &223.44 &	220.47 &	263.28 &	298.47 &	$\colorbox[gray]{0.9}{1161.25}$ &1443.57  \\\midrule
\multirow{3}{*}{8} &ROD  & 238.68 &223.88 &221.36 &263.24 &	301.40 &	1260.10 &1583.87  \\
&MOD  & 239.13 &225.16 &	222.60 &	264.75 &	297.75 &	1170.28 	&1464.51  \\
&EOD  & $\colorbox[gray]{0.9}{234.94}$ &$\colorbox[gray]{0.9}{221.31}$ &$\colorbox[gray]{0.9}{218.57} $&	$\colorbox[gray]{0.9}{261.56}$ &	$\colorbox[gray]{0.9}{298.18}$ &	1214.29 &	$\colorbox[gray]{0.9}{1354.92 }$ \\\bottomrule
\end{tabular}
\end{table}

\subsection{DataLight Network Details}
\label{apxsec:network}

The detailed network of DataLight is descriebd as follows, which is consists of four modules:
\begin{itemize}[leftmargin=*]
\item \textbf{Spatial encoding.} The state inputs undergo an initial embedding layer and are then concatenated based on their spatial correlation, as follows:
\begin{align}
\mathbf{F}_{l,i} = \operatorname{Emb}(s_{l, i})\oplus \operatorname{Emb}(s_{l, i+8})\oplus \operatorname{Emb}(s_{l, i+16}),    
\end{align}
where $i$ ranges from $0$ to $7$, corresponding to the 8 segments, and $\oplus$ denotes the concatenate operation. Then, multi-head self-attention (MHA)~\cite{attention} is employed for SE.
\begin{align}
\mathbf{F}_l = \operatorname{MHA}\left (\mathbf{F}_{l,0} \oplus \mathbf{F}_{l,1} \oplus \cdots \oplus\mathbf{F}_{l,7} \right),
\end{align}
where positional encoding is utilized for SE.
\item \textbf{Feature fusion.} Multi-head self-attention (MHA) is  employed to integrate features from all incoming lanes relevant to each phase:
\begin{equation}
\mathbf{F}^p = \operatorname{Mean}(\operatorname{MHA}(\mathbf{F}_l\oplus \mathbf{F}_k), l, k\in \mathcal{L}_p,
\end{equation}
where $p$ denotes each phase, $\mathcal{L}_p$ is the set of participating entering lanes of phase $p$.
\item \textbf{Phase correlation.} The Fused phase feature is processed using MHA to capture inter-phase correlations:
    \begin{equation}
        \mathbf{F}^{\mathcal{P}} = \operatorname{MHA}(\mathbf{F}^{A} \oplus \mathbf{F}^{B} \oplus \mathbf{F}^{C} \oplus \mathbf{F}^{D} ), \mathcal{P} = \{A,B,C,D\},
    \end{equation}
\item \textbf{Q-value prediction.} The phase features are further processed to obtain the Q-value of each action:
\begin{equation}
\widetilde q_d = \operatorname{DuelingBlock}(\operatorname{Embed}(\mathbf{F}^{\mathcal{P}})).
\end{equation} 
where $\operatorname{DuelingBlock}$ is the dueling module~\cite{dueling}.
\end{itemize}

%===============================
\subsection{Details of offline datasets generation}
\label{apxsec:data}

Offline datasets can be generated with different action durations, Table~\ref{tab:od} describes the specific performance of the methods which are used for generating the offline dataset. Apart from COD, each dataset contains 100K transitions. For the COD generated with the action duration as 10s, there is a total of 24480 ($360\times12\times3+360\times16\times2$) transitions;  for the COD generated with the action duration as 15s, there is a total of 16320 ($240\times12\times3+240\times16\times2$) transitions; while for the COD generated with the action duration as 20s, there is a total of 12240 ($180\times12\times3+180\times16\times2$) transitions.

\begin{table*}[ht]
\caption{Specific methods performance used for generating the offline dataset (ATT in seconds).}
\label{tab:od}
\centering
\begin{tabular}{lcccccc}\toprule
\multirow{2}{*}{Dataset } &\multirow{2}{*}{ Action Duraiton} &\multicolumn{3}{c}{ JN } & \multicolumn{2}{c}{ HZ }  \\
\cmidrule { 3 - 7 } & &JN1 & JN2 & JN3 & HZ1 & HZ2 \\\midrule
\multirow{3}{*}{ROD} & 10s & 590.92 &	537.05 &	539.29 &	577.24 &	483.43 \\
& 15s & 559.76& 	507.28 &	515.84 &	555.41 	&464.70\\
& 20s & 560.02 &	514.37 &	519.54 &	572.95 &	471.69 \\\midrule
\multirow{3}{*}{MOD} & 10s & 321.36 &	297.76 &	284.39 &	308.15 &	381.81 \\
& 15s & 289.06 &	262.40 &	259.93 &	308.16& 	345.31 \\
& 20s & 282.29 &	269.97& 	268.85& 	313.74& 	341.57 \\\midrule
\multirow{3}{*}{EOD} & 10s & 247.20 &	226.13& 	223.62& 	264.25& 	325.94 \\
& 15s & 246.76 &	233.85& 	230.41 &	271.68& 	310.25 \\
& 20s & 254.39 &	244.20 &	239.19& 	282.08& 	315.91 \\\midrule
\multirow{3}{*}{COD} & 10s & 612.36 	&549.21 &	542.83 &	554.73 &	462.62 \\
& 15s & 429.27 &	370.34 &	384.89& 	497.87 &	408.31 \\
& 20s & 443.08 &	369.91 &	390.17& 	483.38 &	404.49 \\
\bottomrule
\end{tabular}
\end{table*}

%===============================
\subsection{Model performance with different offline datasets}
\label{apxsec:model_performance}
DataLight is evaluated when trained with different offline datasets. Table~\ref{apxtab:perfor} demonstrates the model performance with ATT in seconds. When trained with ROD, MOD, and EOD, DataLight performs best under JN and HZ with the action duration as 10s (which is used to generate the offline datasets). When trained with COD, DataLight performs best among all the dataset with the action duration as 20s. When tested under NY, DataLight consistently performs best with the action duration as 20s.

\begin{table}[ht]
\caption{DataLight performance under different offline datasets (ATT in seconds).}
\label{apxtab:perfor}
\centering
\begin{tabular}{lcccccccc}\toprule
\multirow{2}{*}{Dataset } &\multirow{2}{*}{ Action Duraiton} &\multicolumn{3}{c}{ JN } & \multicolumn{2}{c}{ HZ }  & \multicolumn{2}{c}{ NY } \\
\cmidrule { 3 - 9 } & &JN1 & JN2 & JN3 & HZ1 & HZ2& NY1 & NY2 \\\midrule
\multirow{3}{*}{ROD} & 10s & 238.68 &	223.88 &	221.36 &	263.24 &	301.40 &	1260.10 	&1583.87 \\
& 15s & 244.43 &	231.72 &	227.91 &	270.56& 	295.92 &	1194.52 &	1513.24 \\
& 20s & 250.33 &	243.16 	&236.95 &	281.64 &	308.50 &	1095.40 	&1367.04 \\\midrule
\multirow{3}{*}{MOD} & 10s & 239.13 &	225.16 &	222.60 &	264.75 &	297.75 &	1170.28 &	1464.51 \\
& 15s & 242.51 &	231.51 &	227.49 &	271.00 &	290.75 &	1118.07 &	1467.25 \\
& 20s & 250.69 &	243.06 &	236.15 &	284.23 &	297.29 &	920.04 	&1221.33  \\\midrule 
\multirow{3}{*}{EOD} & 10s &234.94 &	221.31 &	218.57 	&261.56 &	298.18 &	1214.29 	&1354.92 \\
& 15s &241.01 &	229.68 &	225.80 &	268.67 &	299.68& 	1216.64 	&1527.65 \\
& 20s & 249.01 &	240.34 &	234.67 &	278.76 &	302.59 &	1075.23 &	1393.08 \\\midrule
\multirow{3}{*}{COD} & 10s & 324.98 &	244.20& 	273.76 &	294.08 &	332.12 &	1040.38 &	1427.68 \\
& 15s &271.09 &	252.89 	&248.89 &	296.17 &	326.92 &	1010.09 	&1359.39 \\
& 20s &266.68 &259.53 &	251.35 &	300.67 &	319.80 &	963.13 &	1344.77 \\\bottomrule
\end{tabular}
\end{table}

%===============================
\subsection{Analysis of Decision Transformer}
\label{apxsec:DT}
We realize Decision Transformer (DT)~\cite{DT} for TSC according to the original article. All the offline datasets are generated with action duration as 10s. Table~\ref{apxtab:DT} demonstrates the model performance of DT with different configurations. K indicates the sequence length when modeling DT. The returns-to-go is set as -351 according to the gather the trajectory.
DT performs better under K=1 than K=2, indicating that the sequential modeling cannot advance the model performance.
Furthermore, DT performs better when removing the transformer, indicating modeling returns-to-go, states and action with transformer cannot work.
Overall, we conclude that DT cannot be applied for TSC. And the possible reasons include: 1) the special environment of TSC which the long action duration makes the sequential modeling captures the temporal relations hard; 2) the offline datasets are mixed with trajectories from multiple agents, which may make DT agents confused.

\begin{table}[ht]
\caption{The performance of DT under different configurations (ATT in seconds).}
\label{apxtab:DT}
\centering
\begin{tabular}{lcccccc}\toprule
\multirow{2}{*}{Configure } &\multirow{2}{*}{ Dataset} &\multicolumn{3}{c}{ JN } & \multicolumn{2}{c}{ HZ }  \\
\cmidrule { 3 - 7 } & &JN1 & JN2 & JN3 & HZ1 & HZ2 \\\midrule 
\multirow{3}{*}{K=1 } & ROD &  779.63&	809.38&	722.30&	514.93&	640.53  \\
& MOD & 851.77&	828.99&	796.33&	628.81&	792.93\\
& EOD &  1295.69&	1387.51&	1277.11&	1173.84&	825.32\\\midrule
\multirow{3}{*}{K=2 } & ROD &  997.04&	1004.47&	948.26&	833.25&	630.32 \\
& MOD & 1278.78&	1366.18&	1258.05&	1297.03&	893.14\\
& EOD & 1225.71&	1304.54&	1189.28&	1185.02&	808.74 \\\midrule
\multirow{3}{*}{w/o transformer } & ROD &  621.98&	663.42&	617.96&	610.07&	491.94  \\
& MOD &679.45&	626.30&	620.78&	604.61&	501.80 \\
& EOD & 898.23&	955.31&	866.42&	873.97&	626.35 \\\bottomrule
\end{tabular}
\end{table}

%%%%%%%%%%%%%%%%%%%%%%%%%%%%%%%%%%%%%%%%%%%%%%%%%%%%%%%%%%%%%%%%%%%%%%%%%%%%%%%
\section{Details of Experimental Results}
\label{apxsec:exp}
\begin{table}[ht]
\centering
\caption{DataLight performance with and without our proposed state representations (ATT in seconds).}
\label{apxtab:states}
\begin{tabular}{lccc|cc}\toprule
\multirow{2}{*}{ Method } & \multicolumn{3}{c}{ JN } & \multicolumn{2}{c}{ HZ } \\
\cmidrule { 2 - 6 } & JN1 & JN2 & JN3 & HZ1 & HZ2 \\\midrule
w/o our states & 244.32 &	226.82 &	224.42& 	266.91 &	303.18 \\
DataLight(ROD) &$\colorbox[gray]{0.9}{238.68}$ &	$\colorbox[gray]{0.9}{223.88}$ &	$\colorbox[gray]{0.9}{221.36}$ 	&$\colorbox[gray]{0.9}{263.24}$ &	$\colorbox[gray]{0.9}{301.40}$  \\
Improvement & $\uparrow$ 2.31\%& $\uparrow$ 1.30\%&$\uparrow$  1.36\%&$\uparrow$ 1.37\%&$\uparrow$ 0.59\%\\\midrule
w/o our states & 244.12 &	228.50 &	226.49 &	269.55 &	312.88 \\
DataLight(MOD) & $\colorbox[gray]{0.9}{239.13}$& 	$\colorbox[gray]{0.9}{225.16}$ &	$\colorbox[gray]{0.9}{222.60}$ &	$\colorbox[gray]{0.9}{264.75}$ 	&$\colorbox[gray]{0.9}{297.75}$ \\
Improvement & $\uparrow$ 2.04\%&$\uparrow$ 1.46\%& $\uparrow$ 1.72\% &$\uparrow$ 1.78\% &$\uparrow$ 4.84\%\\\midrule
w/o our states & 239.75 &	224.76 &	222.44 &	264.03 &	300.33 \\
DataLight(EOD) & $\colorbox[gray]{0.9}{234.94}$ &	$\colorbox[gray]{0.9}{221.31}$ &	$\colorbox[gray]{0.9}{218.57}$ &	$\colorbox[gray]{0.9}{261.56}$ &	$\colorbox[gray]{0.9}{298.18}$ \\
Improvement & $\uparrow$ 2.01\%&$\uparrow$ 1.53\%& $\uparrow$ 1.74\%&$\uparrow$ 0.94\%&$\uparrow$ 0.72\%\\\bottomrule
\end{tabular}
\end{table}

%%%%%%%%%%%%%%%%%%%%%%%%%%%%%%%%%%%%%%%%
\begin{table}[t]
\centering
\caption{DataLight performance with and without our proposed reward function (ATT in seconds).}
\label{apxtab:reward}
\begin{tabular}{lccc|cc}\toprule
\multirow{2}{*}{ Method } & \multicolumn{3}{c}{ JN } & \multicolumn{2}{c}{ HZ } \\
\cmidrule { 2 - 6 } & JN1 & JN2 & JN3 & HZ1 & HZ2 \\\midrule
pressure & 369.63 &	373.04 &	359.65& 	312.66 &	355.23 \\
queue length &251.48 	&230.65 &	230.46 &	268.91 &	312.07  \\
DataLight(ROD) & $\colorbox[gray]{0.9}{238.68}$ &	$\colorbox[gray]{0.9}{223.88}$ &	$\colorbox[gray]{0.9}{221.36}$& 	$\colorbox[gray]{0.9}{263.24}$ &	$\colorbox[gray]{0.9}{301.40}$ \\
Improvement & $\uparrow$ 5.09\%&$\uparrow$ 2.94\%& $\uparrow$ 3.95\%&$\uparrow$ 2.11\%&$\uparrow$ 3.42\%\\\midrule
pressure & 269.00 &	238.42 &	236.78 &	278.23& 	318.31 \\
queue length & 243.48 &	225.68 &	224.45 &	$\colorbox[gray]{0.9}{264.60}$& 	305.17 \\
DataLight(MOD) &$\colorbox[gray]{0.9}{239.13}$ &	$\colorbox[gray]{0.9}{225.16}$ &	$\colorbox[gray]{0.9}{222.60}$ &	264.75 	&$\colorbox[gray]{0.9}{297.75}$ \\
Improvement & $\uparrow$ 1.79\%&$\uparrow$ 0.23\%& $\uparrow$ 0.82\%&$\downarrow$ 0.06\%&$\uparrow$ 2.43\%\\\midrule
pressure & 259.16 &	230.19& 	231.89 &	267.77& 	311.20 \\
queue length &238.24 &	222.70 &	220.07 &	262.36 &	299.15 \\
DataLight(EOD) & $\colorbox[gray]{0.9}{234.94 }$&	$\colorbox[gray]{0.9}{221.31}$ &	$\colorbox[gray]{0.9}{218.57}$ &	$\colorbox[gray]{0.9}{261.56}$& $\colorbox[gray]{0.9}{298.18}$ \\
Improvement & $\uparrow$ 1.39\%&$\uparrow$ 0.62\%& $\uparrow$ 0.68\%&$\uparrow$ 0.30\%&$\uparrow$ 0.32\%\\\bottomrule
\end{tabular}
%\end{table}

%%%%%%%%%%%%%%%%%%%%%%%%%%%%%%%%%%%%%%%
%\begin{table}[t]
\centering
\caption{DataLight performance with and without spatial sequential encoding (ATT in seconds).}
\label{apxtab:STE}
\begin{tabular}{lccc|cc}\toprule
\multirow{2}{*}{ Method } & \multicolumn{3}{c}{ JN } & \multicolumn{2}{c}{ HZ } \\
\cmidrule { 2 - 6 } & JN1 & JN2 & JN3 & HZ1 & HZ2 \\\midrule
w/o SE & 242.15 &	225.53& 	224.45 &	264.26 	&$\colorbox[gray]{0.9}{296.99}$ \\
DataLight(ROD) & $\colorbox[gray]{0.9}{238.68}$ &	$\colorbox[gray]{0.9}{223.88} $&	$\colorbox[gray]{0.9}{221.36}$ &	$\colorbox[gray]{0.9}{263.24}$ 	&301.40 \\
Improvement & $\uparrow$ 1.43\%&$\uparrow$ 0.72\%& $\uparrow$ 1.38\%&$\uparrow$ 0.39\%&$\downarrow$ 1.46\% \\\midrule
w/o SE & 240.02 &	225.16 	&223.43 &	$\colorbox[gray]{0.9}{264.69}$ &	303.55 \\
DataLight(MOD) & $\colorbox[gray]{0.9}{239.13} $& 	225.16 &	$\colorbox[gray]{0.9}{222.60} $	&264.75 &	$\colorbox[gray]{0.9}{297.75}$ \\
Improvement & $\uparrow$ 0.37\%&$\uparrow$ 0.00\%& $\uparrow$ 0.37\%&$\downarrow$ 0.02\%&$\uparrow$ 1.91\%\\\midrule
w/o SE & 236.51 &	221.93 &	219.55 &	262.29 &	301.83 \\
DataLight(EOD) & $\colorbox[gray]{0.9}{234.94}$ &	$\colorbox[gray]{0.9}{221.31}$& 	$\colorbox[gray]{0.9}{218.57}$& 	$\colorbox[gray]{0.9}{261.56} $&	$\colorbox[gray]{0.9}{298.18}$ \\
Improvement & $\uparrow$ 0.66\%&$\uparrow$ 0.28\%& $\uparrow$ 0.45\%&$\uparrow$ 0.28\%&$\uparrow$ 1.21\%\\\bottomrule
\end{tabular}
%\end{table}

%%%%%%%%%%%%%%%%%%%%%%%%%%%%%%%%%%%%%%%

%\begin{table}[t]
\centering
\caption{Performance comparison under cyclical offline dataset(ATT in secodns).}
\label{apxtab:COD}
\begin{tabular}{lccc|cc|cc}\toprule
\multirow{2}{*}{ Method } & \multicolumn{3}{c}{ JN } & \multicolumn{2}{c}{ HZ } & \multicolumn{2}{c}{ NY } \\
\cmidrule { 2 - 8 } & JN1 & JN2 & JN3 & HZ1 & HZ2& NY1& NY2 \\\midrule 
BC & 607.36 &	365.27 &	385.22& 	476.28 &	398.54 	&1457.91 &	1715.38 \\
BCQ & 431.22 &	378.34 &	392.17 &	372.35 &	362.48 &	1313.79 &	1651.10 \\
CQL & 431.22 &	378.34 	&392.17 &	372.35 &	362.48 &	1313.79 	&1651.10 \\\midrule
\textbf{DataLight} &$\colorbox[gray]{0.9}{266.68}$ &	$\colorbox[gray]{0.9}{259.53}$ 	&$\colorbox[gray]{0.9}{251.35}$ &$\colorbox[gray]{0.9}{300.67}$ &	$\colorbox[gray]{0.9}{319.80}$ &	$\colorbox[gray]{0.9}{963.13}$ &	$\colorbox[gray]{0.9}{1344.77}$  \\
Improvement & $\uparrow$ 10.65\%& $\uparrow$8.50\% &$\uparrow$ 8.90\%& $\uparrow$ 13.18\% & $\uparrow$ 10.83\%& $\uparrow$ 26.69\% &$\uparrow$ 17.65\%\\\bottomrule
\end{tabular}
%\end{table}

%%%%%%%%%%%%%%%%%%%%%%%%%%%%%%%%%%%%%%%
%\begin{table}[t]
\centering
\caption{DataLight performance under different action durations when trained with mixed COD (ATT in seconds)}
\label{apxtab:CODm}
\begin{tabular}{lccc|cc|cc}\toprule
\multirow{2}{*}{ Action duration } & \multicolumn{3}{c}{ JN } & \multicolumn{2}{c}{ HZ } & \multicolumn{2}{c}{ NY } \\
\cmidrule { 2 - 8 } & JN1 & JN2 & JN3 & HZ1 & HZ2& NY1& NY2 \\\midrule 
10s &  264.74 &$	\colorbox[gray]{0.9}{229.83}$ &	233.15 &	$\colorbox[gray]{0.9}{266.25}$ &	305.47 &	978.94 	&1258.38 \\
15s &  $\colorbox[gray]{0.9}{250.19}$ &	232.40 	&$\colorbox[gray]{0.9}{228.76} $&	271.46 &	$\colorbox[gray]{0.9}{299.79}$& 	959.41 &	1268.76 \\
20s &  253.12 &	243.04 &	237.47 &	281.09 &	303.46 &	969.19 &	1282.31 \\
25s & 263.65 &	255.19& 	248.73 &	292.70 &	310.98 &	$\colorbox[gray]{0.9}{922.30}$ &	$\colorbox[gray]{0.9}{1259.75}$ \\
\bottomrule
\end{tabular}
\end{table}

%%%%%%%%%%%%%%%%%%%%%%%%%%%%%%%%%%%%%%%%%%%%%%%%%%%%%%%%%%%%%%%%%%%%%%%%%%%%%%%
\end{appendices}
\end{document}